\documentclass[nohyperref]{article}

\usepackage{microtype}
\usepackage{graphicx}
\usepackage{subfigure}
\usepackage{booktabs} % for professional tables

\usepackage{hyperref}

\usepackage[accepted]{icml2022}

\usepackage{amsmath}
\usepackage{amssymb}
\usepackage{mathtools}
\usepackage{amsthm}
\usepackage{multirow}
\usepackage{arydshln}

\usepackage[capitalize,noabbrev]{cleveref}

\theoremstyle{plain}

\theoremstyle{definition}

\theoremstyle{remark}

\usepackage[textsize=tiny]{todonotes}
\usepackage{algorithm}
\usepackage{algorithmic}
\renewcommand{\algorithmicrequire}{ \textbf{Input:}} %Use Input in the format of Algorithm
\renewcommand{\algorithmicensure}{ \textbf{Output:}}

\icmltitlerunning{Improving Adversarial Robustness via Mutual Information Estimation}

\begin{document}

\twocolumn[
\icmltitle{Improving Adversarial Robustness via Mutual Information Estimation}

% It is OKAY to include author information, even for blind
% submissions: the style file will automatically remove it for you
% unless you've provided the [accepted] option to the icml2022
% package.

% List of affiliations: The first argument should be a (short)
% identifier you will use later to specify author affiliations
% Academic affiliations should list Department, University, City, Region, Country
% Industry affiliations should list Company, City, Region, Country

% You can specify symbols, otherwise they are numbered in order.
% Ideally, you should not use this facility. Affiliations will be numbered
% in order of appearance and this is the preferred way.
\icmlsetsymbol{equal}{*}
\icmlsetsymbol{correspond}{$\dagger$}

\begin{icmlauthorlist}
\icmlauthor{Dawei Zhou}{xd}
\icmlauthor{Nannan Wang}{xd,correspond}
\icmlauthor{Xinbo Gao}{cu}
\icmlauthor{Bo Han}{hkbu}
\icmlauthor{Xiaoyu Wang}{cuhk}
\icmlauthor{Yibing Zhan}{jd}
\icmlauthor{Tongliang Liu}{tml}

\end{icmlauthorlist}

\icmlaffiliation{xd}{State Key Laboratory of Integrated Services Networks, School of Telecommunications Engineering, Xidian University}
\icmlaffiliation{hkbu}{Department of Computer Science, Hong Kong Baptist University}
\icmlaffiliation{cu}{Chongqing Key Laboratory of Image Cognition, Chongqing University of Posts and Telecommunications}
\icmlaffiliation{tml}{TML Lab, Sydney AI Centre, The University of Sydney}
\icmlaffiliation{cuhk}{The Chinese University of Hong Kong (Shenzhen)}
\icmlaffiliation{jd}{JD Explore Academy}

\icmlcorrespondingauthor{Nannan Wang}{nnwang@xidian.edu.cn}

% You may provide any keywords that you
% find helpful for describing your paper; these are used to populate
% the "keywords" metadata in the PDF but will not be shown in the document
\icmlkeywords{Machine Learning, ICML}

\vskip 0.3in
]

% this must go after the closing bracket ] following \twocolumn[ ...

% This command actually creates the footnote in the first column
% listing the affiliations and the copyright notice.
% The command takes one argument, which is text to display at the start of the footnote.
% The \icmlEqualContribution command is standard text for equal contribution.
% Remove it (just {}) if you do not need this facility.

\printAffiliationsAndNotice{}  % leave blank if no need to mention equal contribution
% \printAffiliationsAndNotice{\icmlEqualContribution} % otherwise use the standard text.

\begin{abstract}
Deep neural networks (DNNs) are found to be vulnerable to adversarial noise. They are typically misled by adversarial samples to make wrong predictions.
To alleviate this negative effect, in this paper, we investigate the dependence between outputs of the target model and input adversarial samples from the perspective of information theory, and propose an adversarial defense method. Specifically, we first measure the dependence by estimating the mutual information (MI) between outputs and the natural patterns of inputs (called \textit{natural MI}) and MI between outputs and the adversarial patterns of inputs (called \textit{adversarial MI}), respectively. We find that adversarial samples usually have larger adversarial MI and smaller natural MI compared with those \textit{w.r.t.} natural samples. 
%This indicates that the output is closely relevant to the adversarial pattern, while being significantly irrelevant to the natural pattern that contains the information about the true objective.
Motivated by this observation, we propose to enhance the adversarial robustness by maximizing the natural MI and minimizing the adversarial MI during the training process. In this way, the target model is expected to pay more attention to the natural pattern that contains objective semantics. Empirical evaluations demonstrate that our method could effectively improve the adversarial accuracy against multiple attacks. 
\end{abstract}

\section{Introduction}
\label{section1}
\footnote{This version corrects an error in the published version.}Deep neural networks (DNNs) have been demonstrated to be vulnerable to adversarial examples \cite{szegedy2013intriguing, goodfellow2014explaining,liao2018defense,shen2017ape,ma2018characterizing,wu2020adversarial, pmlr-v139-zhou21e}. The adversarial samples are typically generated by adding imperceptible but adversarial noise to natural samples. The vulnerability of DNNs seriously threatens many decision-critical deep learning applications, such as image processing \cite{lecun1998gradient, Zagoruyko2016WRN, 2017Mask, xia2020part,  ma2021understanding, xia2021robust}, natural language processing \cite{sutskever2014sequence} and speech recognition \cite{sak2015fast}. The urgent demand to reduce these security risks prompts the development of adversarial defenses.

Many researchers have made extensive efforts to improve the adversarial robustness of DNN. A major class of adversarial defense focuses on exploiting adversarial samples to help train the target model to achieve adversarially robust performance \cite{madry2017towards,ding2019sensitivity,zhang2019theoretically,wang2019improving,zhou2021modeling,zheng2021regularizing,yang2021class}. However, the dependence between the output of the target model and the input adversarial sample has not been well studied yet. Explicitly measuring this dependence could help train the target model to make predictions that are closely relevant to the given objectives \cite{belghazi2018mutual,sanchez2020learning}.

In this paper, we investigate the dependence from the perspective of information theory. Specifically, we exploit the mutual information (MI) to explicitly measure the dependence of the output on the adversarial sample. MI is an entropy-based measure that can reflect the dependence degree between variables. A larger MI typically indicates stronger dependence between the two variables. However, directly exploiting MI between the input and its corresponding output (called \textit{standard MI}) to measure the dependence has limitations in improving classification accuracy for adversarial samples.

\begin{figure}[t]
\begin{center}
\vskip 0.1in
\centerline{\includegraphics[width=3.0in]{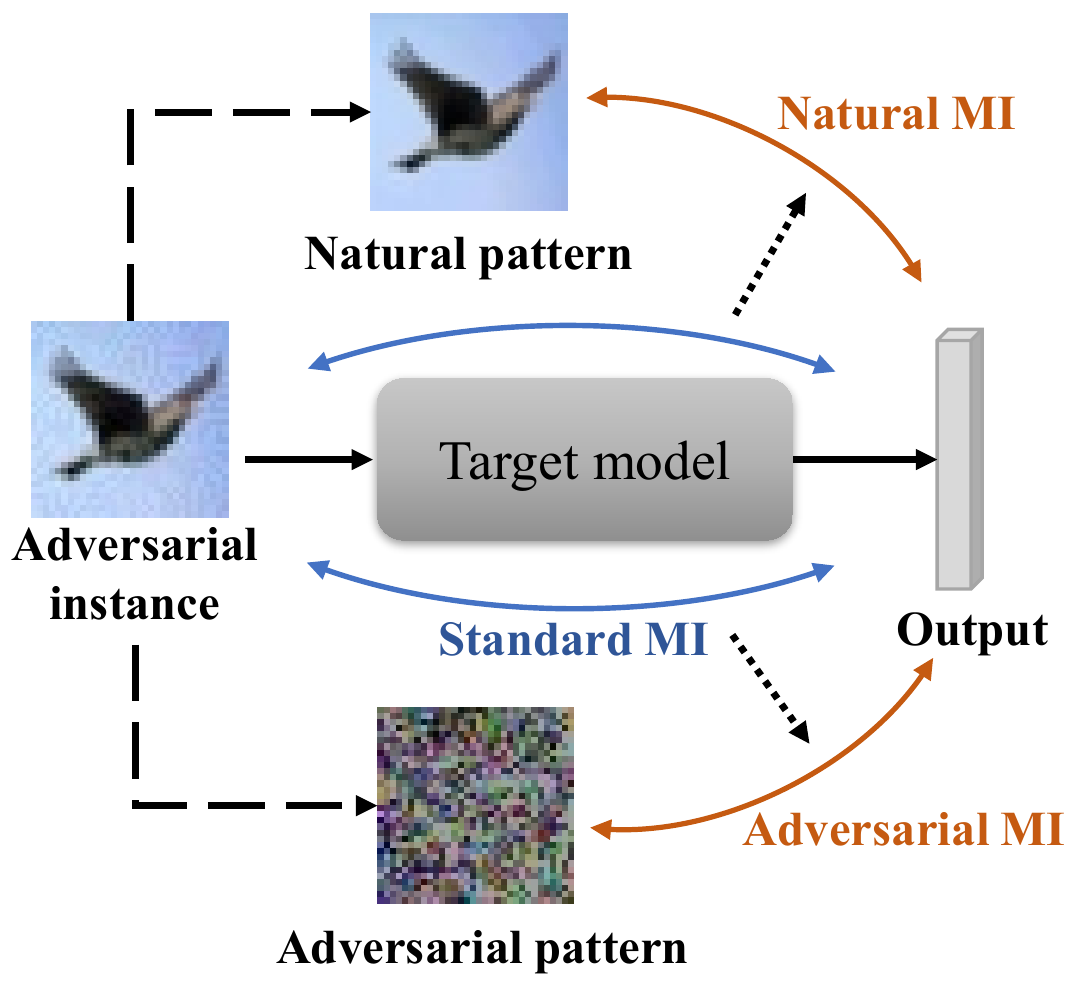}}
\caption{A visual illustration of disentangling the standard MI into the natural MI and the adversarial MI. The \textit{longdash} lines show that the adversarial sample is disassembled into the natural pattern (derived from the natural sample) and the adversarial pattern (derived from the adversarial noise). The \textit{dotted} lines denote the operation of disentangling the standard MI into the natural MI and the adversarial MI.}
\label{fig1}
\end{center}
\vskip -0.3in
\end{figure}

Note that adversarial samples contain natural and adversarial patterns. As shown in \cref{fig1}, given a target model and an adversarial sample, the natural pattern is derived from the corresponding natural sample, and the adversarial pattern is derived from the adversarial noise in the adversarial sample. The adversarial pattern controls the flip of the prediction from the correct label to the wrong label \cite{ilyas2019adversarial}. The standard MI of the adversarial sample reflects the confused dependence of the output on the natural pattern and the adversarial pattern. Maximizing the standard MI of the adversarial sample to guide the target model may result in a larger dependence between the output and the adversarial pattern. This may cause more interference with the prediction of the target model. Therefore, directly maximizing the standard MI to help train the target model cannot surely promote the adversarial robustness.

To handle this issue, we propose to disentangle the standard MI to explicitly measure the dependence of the output on the natural pattern and the adversarial pattern, respectively. As shown in \cref{fig1}, we disentangle the standard MI into the \textit{natural MI} (i.e., the MI between the output and the natural pattern) and the \textit{adversarial MI} (i.e., the MI between the output and the adversarial pattern). To present the reasonability of the disentanglement, we theoretically demonstrate that standard MI is closely related to the linear sum of natural MI and adversarial MI. In addition, how to effectively estimate the natural MI and the adversarial MI is a crucial problem. Inspired by the \textit{mutual information maximization} in \citet{hjelm2018learning,zhu2020learning}, we design a neural network-based method to train two MI estimators to estimate the natural MI and adversarial MI respectively. The detailed discussion can be found in \cref{section3.2}. 

%Moreover, we find that adversarial samples usually have larger adversarial MI and smaller natural MI compared with natural samples. To adequately represent the inherent difference between the MI of natural samples and that of adversarial samples, we design an optimization mechanism for the MI estimation. This is, we minimize the natural MI of the adversarial sample and minimize the adversarial MI of the natural sample during training the estimators. The detailed discussion can be found in \cref{section3.2.1}. 

Based on the above MI estimation, we develop an adversarial defense algorithm called \textit{natural-adversarial mutual information-based defense} (NAMID) to enhance the adversarial robustness. Specifically, we introduce an optimization strategy using the natural MI and the adversarial MI on the basis of the adversarial training manner. The optimization strategy is to maximize the natural MI of the input adversarial sample and minimize its adversarial MI simultaneously. By iteratively executing the procedures of generating adversarial samples and optimizing the model parameters, we can learn an adversarially robust target model. 

The main contributions in this paper are as follows:
\begin{itemize}
    \item Considering the adversarial sample has the natural pattern and the adversarial pattern, we propose the natural MI and the adversarial MI to explicitly measure the dependence of the output on the different patterns.
    \item We design a neural network-based method to effectively estimate the natural MI and the adversarial MI. By exploiting the MI estimation networks, we develop a defense algorithm to train a robust target model. 
    \item We empirically demonstrate the effectiveness of the proposed defense algorithm on improving the classification accuracy. The evaluation experiments are conducted against multiple adversarial attacks.
\end{itemize}

The rest of this paper is organized as follows. In Section~\ref{section2}, we introduce some preliminary information and briefly review related works. In Section~\ref{section3}, we propose the natural MI and the adversarial MI, and develop an adversarial defense method. Experimental results are provided in Section~\ref{section4}. Finally, we conclude this paper in Section~\ref{section5}.

\section{Preliminaries}
\label{section2}
In this section, we introduce some preliminary about notation, the problem setting and mutual information (MI). We also review some representative literature on adversarial attacks and defenses.

\noindent\textbf{Notation.} 
We use \textit{capital} letters such as $X$ and $Y$ to represent random variables, and \textit{lower-case} letters such as $x$ and $y$ to represent realizations of random variables $X$ and $Y$, respectively. For norms, we denote by $\|x\|$ a generic norm, by $\|x\|_{\infty}$ the $L_{\infty}$ norm of $x$, and by $\|x\|_{2}$ the $L_{2}$ norm of $x$. In addition, let $\mathbb{B}(x, \epsilon)$ represent the neighborhood of $x$: $\{\tilde{x}:\|\tilde{x}-x\| \leq \epsilon$\}, where $\epsilon$ is the perturbation budget. We define the \textit{classification function} as $f: \mathcal{X} \rightarrow \{1,2,\ldots,C\}$. It can be parameterized, e.g., by a deep neural network $h_{\theta}$ with model parameter $\theta$.

\noindent\textbf{Problem setting.} 
In this paper, we focuses on a classification task under the adversarial environment. Let $X$ and $Y$ be the variables for natural instances and ground-truth labels respectively. We sample natural data $\{(x_i, y_i)\}_{i=1}^{n}$ according to the distribution of variables $(X,Y)$. Given a pair of natural data $(x,y)$, the adversarial instance $\tilde{x}$ satisfies the following constraint:
\begin{equation}
\label{eq1}
f\left(\tilde{x}\right) \neq y \quad \text { s.t.} \quad\left\|x-\tilde{x}\right\| \leq \epsilon \text{,}
\end{equation}
where $\tilde{x}=x+n$, $n$ denotes the adversarial noise. Our aim is to develop an adversarial defense method to help train the classification model $h_{\theta}$ to make normal predictions.

\noindent\textbf{Mutual information.}
MI is an entropy-based measure that can reflect the dependence degree between variables. A larger MI typically indicates a stronger dependence between the two variables. Various methods have been proposed for estimating MI \cite{moon1995estimation,darbellay1999estimation,kandasamy2015nonparametric,moon2017ensemble}. A representative and efficient estimator is the mutual information neural estimator (MINE) \cite{belghazi2018mutual}. MINE is empirically demonstrated its superiority in estimation accuracy and efficiency, and proved that it is strongly consistent with the true MI. Besides, the work in \citet{hjelm2018learning} points that using the complete input to estimate MI is often insufficient for classification task. Instead, estimating MI between the high-level feature and local patches of the input is more suitable. Therefore, in this paper, we refer the local DIM estimator \cite{hjelm2018learning} to estimate MI. The definition of MI and other details are presented in \cref{appendix_1}

\noindent\textbf{Adversarial attacks.} 
Adversarial noise can be crafted by optimization-based attacks, such as PGD \cite{madry2017towards}, AA \cite{croce2020reliable}, CW \cite{carlini2017towards} and DDN \cite{rony2019decoupling}. Besides, some attacks such as FWA \cite{wu2020stronger} and STA \cite{xiao2018spatially} focus on mimicking non-suspicious vandalism by exploiting the geometry and spatial information. These attacks constrain the perturbation boundary by a small norm-ball $\|\cdot\|_{p} \leq \epsilon$, so that their adversarial instances can be perceptually similar to natural instances. 

\noindent\textbf{Adversarial defenses.} The issue of adversarial attacks promotes the development of adversarial defenses. A major class of adversarial defense methods is devoted to enhance the adversarial robustness in an adversarial training manner \cite{madry2017towards,ding2019sensitivity,zhang2019theoretically,wang2019improving}. They augment training data with adversarial samples and use a min-max formulation to train the target model \cite{madry2017towards}. However, these methods do not explicitly measure the dependence between the adversarial sample and its corresponding output. In addition, some data pre-processing based methods try to remove adversarial noise by learning denoising functions \cite{liao2018defense,naseer2020self,zhou2021removing} or feature-squeezing functions \cite{guo2017countering}. However, these methods may suffer from the problems of human-observable loss \cite{xu2017feature} and residual adversarial noise \cite{liao2018defense}, which would affect the final prediction. To avoid the above problems, we propose to exploit the natural MI and the adversarial MI to learn an adversarially robust classification model in the adversarial training manner.
\vskip -0.1in

\section{Methodology}
\label{section3}
In this section, we first illustrate the motivation for using mutual information (MI) and disentangling the standard mutual information into the \textit{natural MI} and the \textit{adversarial MI} (\cref{section3.1}). Next, we theoretically prove the reasonability of the disentanglement and introduce how to effectively estimate the natural MI and the adversarial MI (\cref{section3.2}). Finally, we propose an adversarial defense algorithm which contains an MI-based optimization strategy (\cref{section3.3}). The code is available at \url{https://github.com/dwDavidxd/MIAT}.

\subsection{Motivation}
\label{section3.1}
For adversarial samples, The predictions of the target model are typically significantly irrelevant to the given objectives in the inputs. Studying the dependence between the adversarial sample and its corresponding output is considered to be beneficial for improving the adversarial robustness. The dependence could be exploited as supervision information to help train the target model to make correct predictions.

Estimating the standard MI of the adversarial sample (i.e., the MI between the adversarial sample and its corresponding output in a target model) is a simple strategy to measure the dependence. However, different from natural samples, adversarial samples have two patterns, i.e., the natural pattern and the adversarial pattern. The standard MI cannot respectively consider the dependence of the output on the different patterns, which may limit its performance in helping the target model improve the adversarial robustness. 

\begin{figure}[t]
\begin{center}
\vskip 0.1in
\centerline{\includegraphics[width=2.8in]{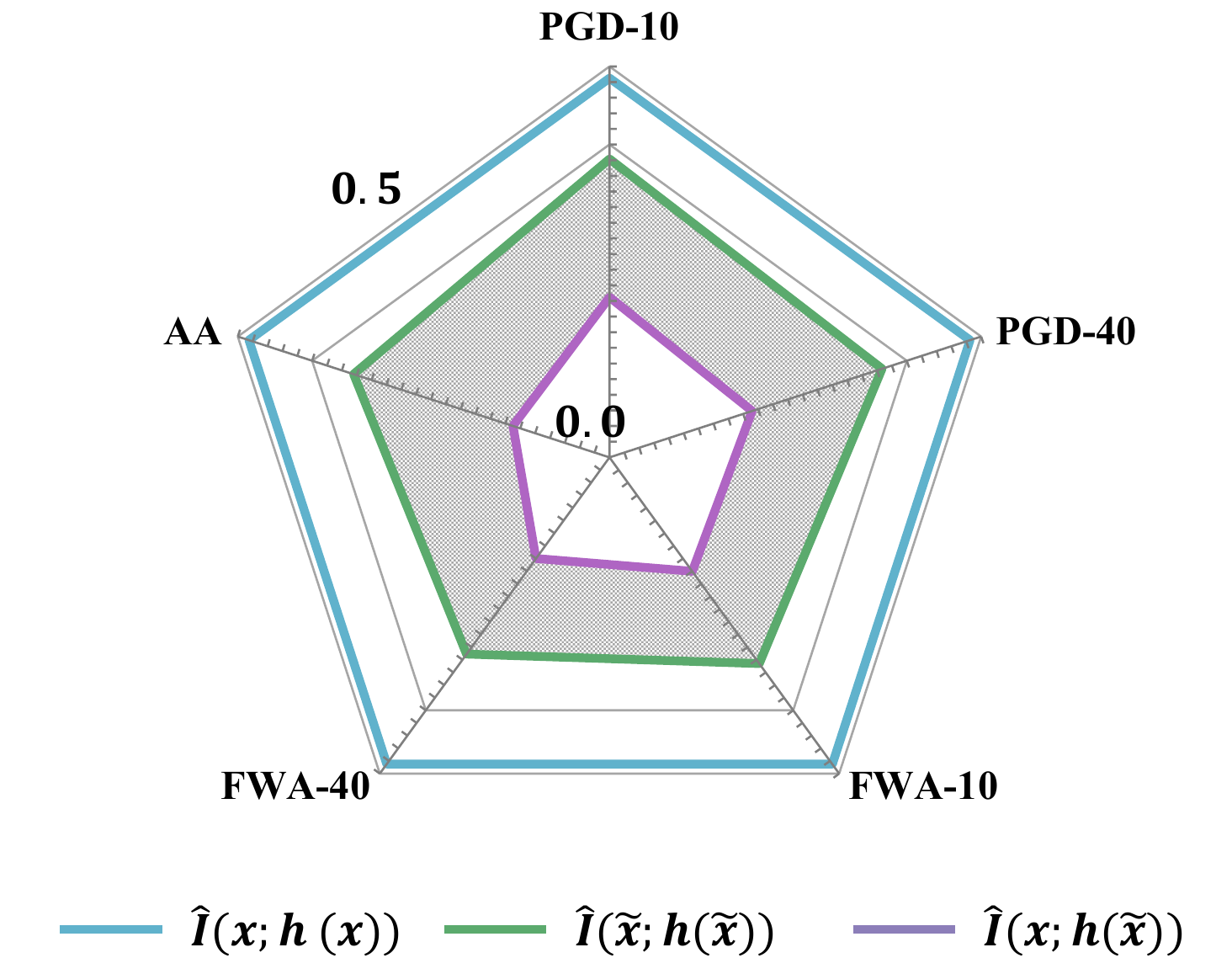}}
\caption{The visualization of the proof-of-concept experiment. Given the natural instance $x$, its adversarial instance $\tilde{x}$ and a target model $h$, the logit output of $h$ for $\tilde{x}$ is denoted by $h(\tilde{x})$. We respectively estimate the standard MI of the natural instance $\widehat{I}(x, h(x))$, the standard MI of the adversarial instance $\widehat{I}(\tilde{x}, h(\tilde{x}))$ and the MI between $h(\tilde{x})$ and $x$. The \textit{shaded} area represents the difference between $\widehat{I}(\tilde{x}, h(\tilde{x}))$ and $\widehat{I}(x, h(\tilde{x}))$.}
\label{fig2}
\vskip -0.3in
\end{center}
\end{figure}

Specifically, the natural pattern is derived from the original natural sample. It provides available information for the target model to produce the right output. The adversarial pattern is derived from the adversarial noise. It controls the flip of the prediction from the correct label to the wrong label \cite{ilyas2019adversarial}. Both the natural and adversarial patterns cause important impacts on the output, but they are mutually exclusive. Therefore, the standard MI actually measures a confused dependence.

To clearly illustrate the confused dependence, we conduct a proof-of-concept experiment. We randomly select a set of natural instances and use five attacks to generative adversarial instances. We use a classification model as the target model. The MI estimator is trained on natural instances and their outputs via the MI maximization \cite{hjelm2018learning}. By exploiting the estimator, we respectively estimate the standard MI of the natural instance, the standard MI of the adversarial instance and the MI between the natural instance and the output for the corresponding adversarial instance. The details of the experiment are presented in \cref{appendix_2}

As shown in \cref{fig2}, the result shows that the standard MI of the adversarial sample is indeed smaller than that of the natural sample. However, it is still significantly larger than the MI between the output and natural pattern only (see the pink line). This shows that the standard MI of the adversarial sample contains the dependence of the output on the adversarial pattern. Thus, maximizing the standard MI may increase the dependence of the output on the adversarial pattern and cause more disturbance to the prediction. Directly maximizing the standard MI cannot comprehensively promote the target model to make more accurate predictions for adversarial samples. 

To solve this problem, in this paper, we propose to disentangle the standard MI into two parts related to the natural and adversarial patterns respectively.

\subsection{Natural MI and adversarial MI}
\label{section3.2}
We define two new concepts: \textit{natural mutual information} (natural MI) and \textit{adversarial mutual information} (adversarial MI). The natural MI is MI between the output and the natural pattern of the input. The adversarial MI is MI between the output and the adversarial pattern of the input. To explicitly measure the dependence of the output on different patterns of the input, we need to disentangle the standard MI into the natural MI and the adversarial MI. 

\subsubsection{Disentangling the standard MI}
\label{section3.2.1}
In this section, we introduce how to disentangle the standard MI and describe the reasonability of the disentanglement. We first provide Theorem 1 to illustrate the transformation relationship of MI among four variables.

\noindent \textbf{Theorem 1.} 
Let $X, \widetilde{X}, N, Z$ denote four random variables respectively, where $\widetilde{X}=X+N$. Let $\widetilde{\mathcal{X}}$ be the feature space of $\widetilde{X}$ and $\mathcal{Z}$ be the feature space of $Z$. 
%$\widetilde{X}$ and $Z$ are random variables that follow the marginal distribution $\mu_{\widetilde{X}}$ and $\mu_{Z}$ respectively. 
Then, for any function $h: \widetilde{\mathcal{X}} \rightarrow \mathcal{Z}$, we have 
\begin{equation}
\label{eq2}
\begin{aligned}
 I(\widetilde{X};Z) &= I(X;Z) + I(N;Z) - I(X;N;Z) \\ &+ H(Z|X,N) - H(Z|\widetilde{X}) \text{,}
\end{aligned}
\end{equation}
where $I(\cdot;\cdot)$ denotes the MI between two variables and $I(\cdot;\cdot;\cdot)$ denotes the MI between two three variables. A detailed proof is provided in \cref{appendix_3}. 

Then, we apply Theorem 1 to the adversarial learning setting and obtain Corollary 1.

\noindent \textbf{Corollary 1.}
Let $X, \widetilde{X}, N$ denote the random variables for the natural instance, adversarial instance and adversarial noise respectively, where $\widetilde{X}=X+N$. Given a function parameterized by a target model $h_{\theta}$ with model parameter $\theta$, the logit output of $h_{\theta}$ for $\widetilde{X}$ is denoted by $h_{\theta}(\widetilde{X})$. Considering that the effects of the natural instance and the adversarial noise on the output are mutually exclusive, we assume that the MI between $X, N$ and $h_{\theta}(\widetilde{X})$ (i.e., $I(X;N;h_{\theta}(\widetilde{X}))$) is small. We also assume that the difference between $H(Z|X,N)$ and $H(Z|\widetilde{X})$ is small (see \cref{appendix_4} for more details), then we have:
\vskip -0.2in
\begin{equation}
\label{eq3}
I(\widetilde{X};h_{\theta}(\widetilde{X})) \approx \underbrace{I(X;h_{\theta}(\widetilde{X}))}_{I_{N}}+\underbrace{I(N;h_{\theta}(\widetilde{X}))}_{I_{A}} \text{,}
\end{equation}
\vskip -0.15in
where $I_{N}$ denote the MI between the output and the natural instance and $I_{A}$ denote the MI between the output and the adversarial noise. 

Actually, we could use the original natural instance and the adversarial noise to represent the natural pattern and the adversarial pattern of the input respectively. In this way, $I_{N}$ and $I_{A}$ could denote the natural MI and the adversarial MI respectively.

According to \cref{eq3}, we can approximately disentangle the standard MI into the natural MI and the adversarial MI. The latter two can not only reflect the dependency between input and output as the former, but also provides independent measurements for different patterns. This is more conducive to designing specific optimization strategies for the two patterns to better alleviate the negative effects of the adversarial noise.

\subsubsection{Estimating the natural MI and the adversarial MI}
\label{section3.2.2}
The local DIM estimation method has been demonstrated to be efficient for estimating MI \cite{hjelm2018learning,zhu2020learning}. We thus first use this method to train a estimation network for the natural MI and the adversarial MI respectively. Let $E_{\phi_{n}}$ denote the estimation network for the natural MI and $E_{\phi_{a}}$ denote the estimation network for the adversarial MI. Considering the inherent close relevance between the natural/adversarial pattern and the output for the natural/adversarial sample, we naturally use the natural/adversarial samples to train the $E_{\phi_{n}}$/$E_{\phi_{a}}$. The optimization goals for $\phi_{n}$ and $\phi_{a}$ are as follows:

\vskip -0.1in
\begin{equation}
\label{eq4}
\begin{aligned}
&\widehat{\phi}_{n}=\underset{\phi_{n} \in \Phi_{N}}{\arg \max } \,E_{\phi_{n}}(h_{\theta_{0}}(X)) \text{,} \\
&\widehat{\phi}_{a}=\underset{\phi_{a} \in \Phi_{A}}{\arg \max } \, E_{\phi_{a}}(h_{\theta_{0}}(\widetilde{X})) \text{,}
\end{aligned}
\end{equation}
\vskip -0.1in

where $\Phi_{N}$ and $\Phi_{A}$ denote the sets of model parameters, $h_{\theta_0}$ denotes the pre-trained target model. It can be naturally trained or be adversarially trained. We use a ResNet-18 optimized by standard AT \cite{madry2017towards} as the pre-trained model $h_{\theta_0}$.  $E_{\widehat{\phi}_{n}}\left(\cdot \right)$ is the estimated natural MI and $E_{\widehat{\phi}_{a}}\left(\cdot \right)$ is the estimated adversarial MI, i.e.,  $E_{\widehat{\phi}_{n}}\left(h_{\theta}(X)\right)=\widehat{I}_{N}(X; h_{\theta}(X))$, $E_{\widehat{\phi}_{n}}(h_{\theta}(\widetilde{X}))=\widehat{I}_{N}(X; h_{\theta}(\widetilde{X}))$ and $E_{\widehat{\phi}_{a}}\left(h_{\theta}(X)\right)=\widehat{I}_{A}(N; h_{\theta}(X))$,  $E_{\widehat{\phi}_{a}}(h_{\theta}(\widetilde{X}))=\widehat{I}_{A}(N; h_{\theta}(\widetilde{X}))$.

By exploiting the two MI estimation network, we estimate the natural MI of the natural sample and the adversarial sample (i.e., $\widehat{I}_{N}(X, h_{\theta_{0}})$ and $\widehat{I}_{N}(\widetilde{X}, h_{\theta_{0}})$), and the adversarial MI of the natural sample and the adversarial sample (i.e., $\widehat{I}_{A}(X, h_{\theta_{0}})$ and $\widehat{I}_{N}(\widetilde{X}, h_{\theta_{0}})$), respectively. We find that adversarial samples usually have larger adversarial MI and smaller natural MI compared with those \textit{w.r.t.} natural samples, which is consistent with our intuitive cognition. However, the change is relatively insignificant, and thus has limitations in reflecting the difference between the adversarial sample and the natural sample in the natural MI and the adversarial MI. We will show this observation later in \cref{section4.2}. 

To adequately represent the inherent difference in the natural MI and the adversarial MI between the natural sample and the adversarial sample, we design an optimization mechanism for the MI estimation. This is, we minimize the natural MI of the adversarial sample and minimize the adversarial MI of the natural sample during training the estimators. In addition, to estimate the two MI more accurately, we select samples that are correctly predicted by the target model and the corresponding adversarial samples are wrongly predicted, to train the estimation networks. The reformulated optimization goals are as follows:
\begin{equation}
\label{eq5}
\begin{aligned}
&\widehat{\phi}_{n}=\underset{\phi_{n} \in \Phi_{N}}{\arg \max } \,[E_{\phi_{n}}( h_{\theta_{0}}(X^{\prime})) - E_{\phi_{n}}(h_{\theta_{0}}(\widetilde{X}^{\prime}))] \text{,} \\
&\widehat{\phi}_{a}=\underset{\phi_{a} \in \Phi_{A}}{\arg \max } \, [E_{\phi_{a}}(h_{\theta_{0}}(\widetilde{X}^{\prime})) - E_{\phi_{a}}\left(h_{\theta_{0}}(X^{\prime}\right))] \text{,}
\end{aligned}
\end{equation}

where $X^{\prime}$ is the selected data:
\vskip -0.2in
\begin{equation}
\label{eq6}
    X^{\prime}=\arg _{X}\left[\delta\left(h_{{\theta}_0}(X)\right)=Y \& \delta(h_{{\theta}_0}(\widetilde{X})) \neq Y\right]  \text{,} 
\end{equation}

where $\delta$ is the operation that transforms the logit output into the prediction label.

\subsection{Adversarial defense algorithm}
\label{section3.3}
Based on the above two MI estimation networks, we develop an adversarial defense algorithm called \textit{natural-adversarial mutual information-based defense} (NAMID) to enhance the adversarial robustness. In this section, we first introduce the natural-adversarial MI-based optimization strategy and then illustrate the training algorithm.

\begin{figure*}[t]
\begin{center}
\vskip 0.1in
\centerline{\includegraphics[width=5.6in]{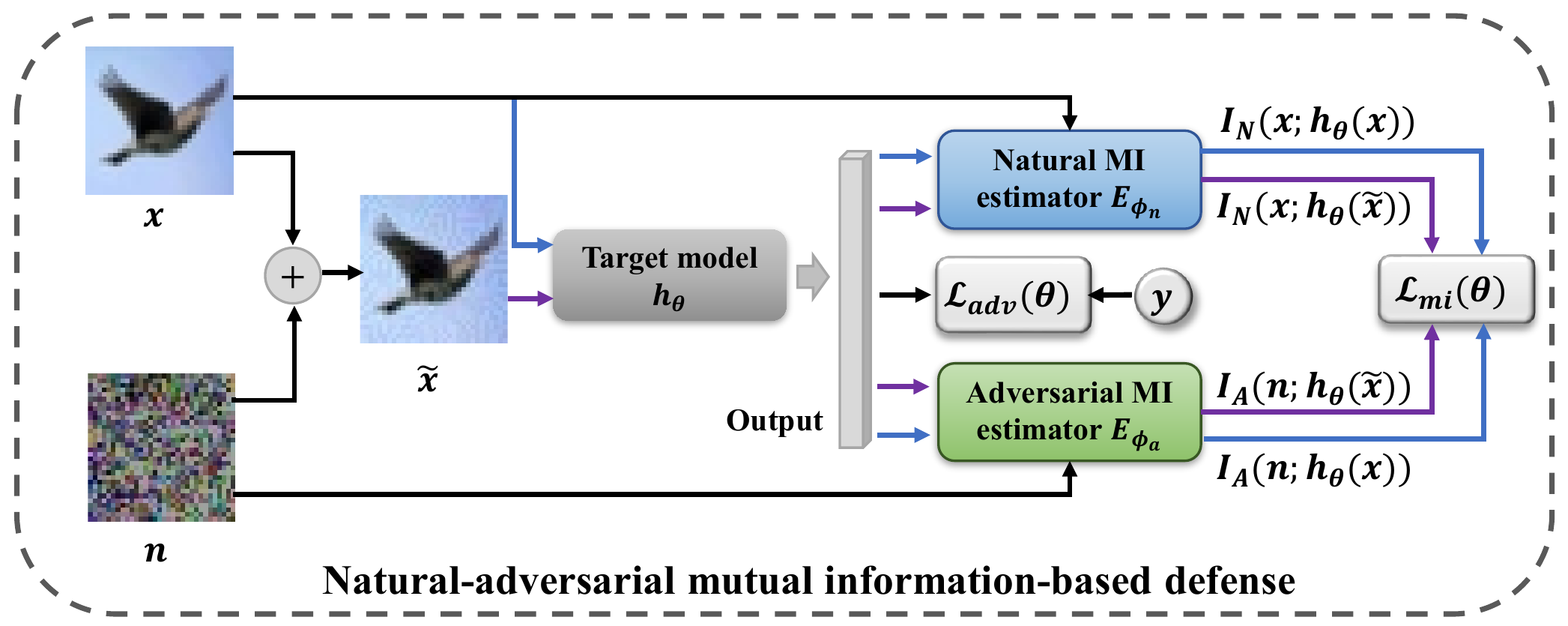}}
\caption{The overview of our proposed NAMID adversarial defense algorithm.}
\label{fig3}
\vskip -0.2in
\end{center}
\end{figure*}

\subsubsection{natural-adversarial MI-based defense}
\label{section3.3.1}
According to the observations and analysis in \cref{section3.1} and \cref{section3.2}, we plan to use the natural MI and the adversarial MI to help train the target model. We aim to guide the target model to increase the attention to the natural pattern while reducing the attention to the adversarial pattern. 

Specifically, The optimization strategy is to maximize the natural MI of the input adversarial sample and minimize its adversarial MI simultaneously. The optimization goal for the target model is as follows:

\begin{equation}
\label{eq7}
\widehat{\theta}=\underset{\theta \in \Theta}{\arg \max }\left[E_{\widehat{\phi}_{n}}(h_{\theta}(\widetilde{X}))-E_{\widehat{\phi}_{a}}(h_{\theta}(\widetilde{X}))\right] \text{.}
\end{equation}

To achieve the optimization goal, we can directly utilize the natural and adversarial MI of the adversarial sample to construct a loss function. However, this loss function does not consider the difference in natural/adversarial MI between the natural sample and the adversarial sample. Thus, we transform this absolute metric-based loss to a relative metric-based loss. In addition, as described in \cref{section3.2.2}, we use the selected samples to compute the loss function. The loss function is formulated as:

\vskip -0.2in
\begin{equation}
\label{eq8}
\begin{gathered}
\mathcal{L}_{m i}(\theta)=\frac{1}{m} \sum_{i=1}^{m}\{\mathcal{L}_{\cos }(E_{\widehat{\phi}_{n}}(h_{\theta}(\tilde{x}_{i}^{\prime})), E_{\widehat{\phi}_{n}}(h_{\theta}(x_{i}^{\prime}))) \\ + \mathcal{L}_{\cos }(E_{\widehat{\phi}_{a}}(h_{\theta}(\tilde{x}_{i}^{\prime}), E_{\widehat{\phi}_{a}}(h_{\theta}(x_{i}^{\prime}))) \\ + \lambda \cdot [E_{\widehat{\phi}_{a}}(h_{\theta}(\tilde{x}_{i}^{\prime}))-E_{\widehat{\phi}_{n}}(h_{\theta}(\tilde{x}_{i}^{\prime}))]\} \text{,}
\end{gathered}
\end{equation}

where $m$ is the number of the selected data $X^{\prime}$ and $\lambda$ is a hyperparameter. $\mathcal{L}_{\cos}(\cdot, \cdot)$ is the cosine similarity-based loss function, i.e., $\mathcal{L}_{\cos}(a, b) = ||1 - sim (a,b)||_1$, $sim(\cdot,\cdot)$ denotes the cosine similarity measure.

The MI-based optimization strategy can exploited together with the adversarial training manner. The overall loss function for training the target model is as follows:

\begin{equation}
\label{eq9}
    \mathcal{L}_{all}(\theta)= \mathcal{L}_{adv}(\theta) + \alpha \cdot \mathcal{L}_{mi}(\theta) \text{,}
\end{equation}
\vskip 0.1in

where $\mathcal{L}_{adv}(\theta)$ is the loss function of the adversarial training method, which is typically the cross-entropy loss between the adversarial outputs and the ground-truth labels: $\mathcal{L}_{adv}(\theta)= -\frac{1}{n} \sum_{i=1}^{n} [\boldsymbol{y_i} \cdot \log(\sigma(h_{\theta}(\tilde{x}_{i})))]$. $n$ is the number of training samples and $\sigma$ denotes the softmax function. $\alpha$ is a trade-off hyperparameter. We provide the overview of the adversarial defense method in \cref{fig3}.
%Since the MI estimator is trained to maximize/minimize the natural/adversarial MI of the natural sample, the hyperparameter $lambda$ in \cref{eq7} is set to a small value. The last item in \cref{eq7} mainly helps to optimize the model parameters during the initial stages of training (i.e. when the accuracy on natural samples is low).

\subsubsection{Training algorithm} 
\label{section3.3.2}
We conduct the adversarial training on the procedures of generating adversarial samples and optimizing the target model parameter. The details of the overall procedure are presented in \cref{alg1}.

 Specifically, the procedure requires the target model $h_{\theta}$ with parameter $\theta$, the natural MI estimation network with parameter $\widehat{\phi}_{n}$, the adversarial MI estimation network with parameter $\widehat{\phi}_{a}$ and perturbation budget $\epsilon$. For the natural instance $x$ in mini-batch $\mathcal{B}=\{x_i\}_{i=1}^{n}$ sampled from natural training set, we first craft adversarial noise $n$ and generate adversarial instance $\tilde{x}$ via the powerful PGD adversarial attack \cite{madry2017towards}. Then, we input the natural and adversarial training data into the target model $h_{\theta}$ and obtain the selected instance $x^{\prime}$, $\tilde{x}^{\prime}$ according to \cref{eq6}. Next, we estimate the natural MI $E_{\widehat{\phi}_{n}}(h_{\theta}(x^{\prime})), E_{\widehat{\phi}_{n}}(h_{\theta}(\tilde{x}^{\prime}))$ and the adversarial MI $ E_{\widehat{\phi}_{a}}(h_{\theta}(x^{\prime})), E_{\widehat{\phi}_{a}}(h_{\theta}(\tilde{x}^{\prime})$. Finally, we optimize the parameter $\theta$ according to \cref{eq9}. By iteratively conduct the adversarial training, $\theta$ is expected to be optimized well.

\begin{algorithm}[t]
   \caption{\small Natural-adversarial mutual information-based defense (NAMID) algorithm}
   \label{alg1}
\begin{algorithmic}[1]
   \begin{small}
   \REQUIRE Target model $h_{\theta}(\cdot)$ parameterized by $\theta$, natural MI estimation network $E_{\widehat{\phi}_{n}}$, adversarial MI estimation network $E_{\widehat{\phi}_{a}}$, batch size $n$, and the perturbation budget $\epsilon$;
   \REPEAT
   \STATE Read mini-batch $\mathcal{B}=\{x_i\}_{i=1}^{n}$ from training set;
   \FOR{$i=1$ to $n$ (in parallel)}
   \STATE Craft adversarial noise $n_{i}$ and generate adversarial instance $\tilde{x}_i$ at the given perturbation budget $\epsilon$ for $x_i$;
   \STATE Forward-pass $x_i$, $\tilde{x}_i$ through $h_{\theta}(\cdot)$ and obtain $h_{\theta}(x_i)$, $h_{\theta}(\tilde{x}_i)$;
   \STATE Select samples according to \cref{eq6};
   \ENDFOR
   \STATE Calculate $\mathcal{L}_{all}$ using \cref{eq9} and optimize $\theta$;
   \UNTIL training converged.
   \end{small}
\end{algorithmic}
\end{algorithm}

\section{Experiments}
\label{section4}
In this section, we first introduce the experiment setups including datasets, attack setting and defense setting in \cref{section4.1}. Then, we show the effectiveness of our optimization mechanism for evaluating MI in \cref{section4.2}. Next, we evaluate the performances of the proposed adversarial defense algorithm in \cref{section4.3}. Finally, we conduct ablation studies in \cref{section4.4}.

\subsection{Experiment setups}
\label{section4.1}
\noindent\textbf{Datasets.}
We verify the effective of our defense algorithm on two popular benchmark datasets, i.e., \textit{CIFAR-10} \cite{krizhevsky2009learning} and \textit{Tiny-ImageNet} \cite{wu2017tiny}. \textit{CIFAR-10} has 10 classes of images including 50,000 training images and 10,000 test images. \textit{Tiny-ImageNet} has 200 classes of images including 100,000 training images, 10,000 validation images and 10,000 test images. Images in the two datasets are all regarded as natural instances. All images are normalized into [0,1], and are performed simple data augmentations in the training process, including random crop and random horizontal flip. 

\noindent\textbf{Model architectures.} We use a ResNet-18 \cite{he2016deep} as the target model for both \textit{CIFAR-10} and \textit{Tiny-ImageNet}. For the MI estimation network, we utilize the same neural network as in \citep{zhu2020learning}. The estimation networks for the natural MI and the adversarial MI have same model architectures.

\noindent\textbf{Baselines.} (1) \textit{Standard AT} \cite{madry2017towards}; (2) TRADES \cite{zhang2019theoretically}; (3) MART \cite{wang2019improving}; and (4)\textit{WMIM}: A defense that refers to \citet{zhu2020learning}. The first three are representative adversarial training methods, and the last one combines adversarial training with standard MI maximization (on adversarial samples).

\noindent\textbf{Attack settings.}
Adversarial data for evaluating defense models are crafted by applying state-of-the-art attacks. These attacks are divided into two categories: $L_{\infty}$-norm attacks and $L_{2}$-norm attacks. The $L_{\infty}$-norm attacks include PGD \cite{madry2017towards}, AA \cite{croce2020reliable}, TI-DIM \cite{dong2019evading,xie2019improving}, and FWA \cite{wu2020stronger}. The $L_{2}$-norm attacks include PGD, CW$_2$ \cite{carlini2017towards} and DDN \cite{rony2019decoupling}. Among them, the AA attack algorithm integrates three non-target attacks and a target attack. Other attack algorithms are utilized as non-target attacks. The iteration number of PGD and FWA is set to 40 with step size $\epsilon/4$. The iteration number of CW$_2$ and DDN are set to 20 respectively with step size 0.01. For \textit{CIFAR-10} and \textit{Tiny-ImageNet}, the perturbation budgets for $L_{2}$-norm attacks and $L_{\infty}$-norm attacks are $\epsilon=0.5$ and $8/255$ respectively.  

\noindent\textbf{Defense settings.} For both \textit{CIFAR-10} and \textit{Tiny-ImageNet}, the adversarial training data for $L_{\infty}$-norm and $L_{2}$-norm is generated by using $L_{\infty}$-norm PGD-10 and $L_{2}$-norm PGD-10 respectively. The step size is $\epsilon/4$ and the perturbation budget is $8/255$ and $0.5$ respectively. The epoch number is set to 100. For fair comparisons, all the methods are trained using SGD with momentum 0.9, weight decay $2 \times 10^{-4}$, batch-size 1024 and an initial learning rate of 0.1, which is divided by 10 at the 75-th and 90-th epoch. In addition, we adjust the hyperparameter settings of the defense methods so that the natural accuracy is not severely compromised and then compare the adversarial accuracy. We set $\alpha=5, \lambda= 0.1$ for our algorithm. 
%We set $\lambda= 5$ for TRADES and MART, and set $\alpha=5, \lambda= 0.1$ for our algorithm. 

\subsection{Effectiveness of MI estimation networks}
\label{section4.2}
In \cref{section3.2.2}, we point out that training the MI estimation network directly by MI maximization may not clearly reflect the difference between the adversarial sample and the natural sample in natural MI and adversarial MI. We thus design an optimization mechanism for training the MI estimation network. To demonstrate the effectiveness of the optimization mechanism, we compare the performance of the estimation networks trained by \cref{eq4} and \cref{eq5} in \cref{fig4}. The performances of the defenses based on the two different estimators are shown in \cref{appendix_5_1}

\begin{figure}[t]
\begin{center}
\vskip 0.1in
\centerline{\includegraphics[width=2.5 in]{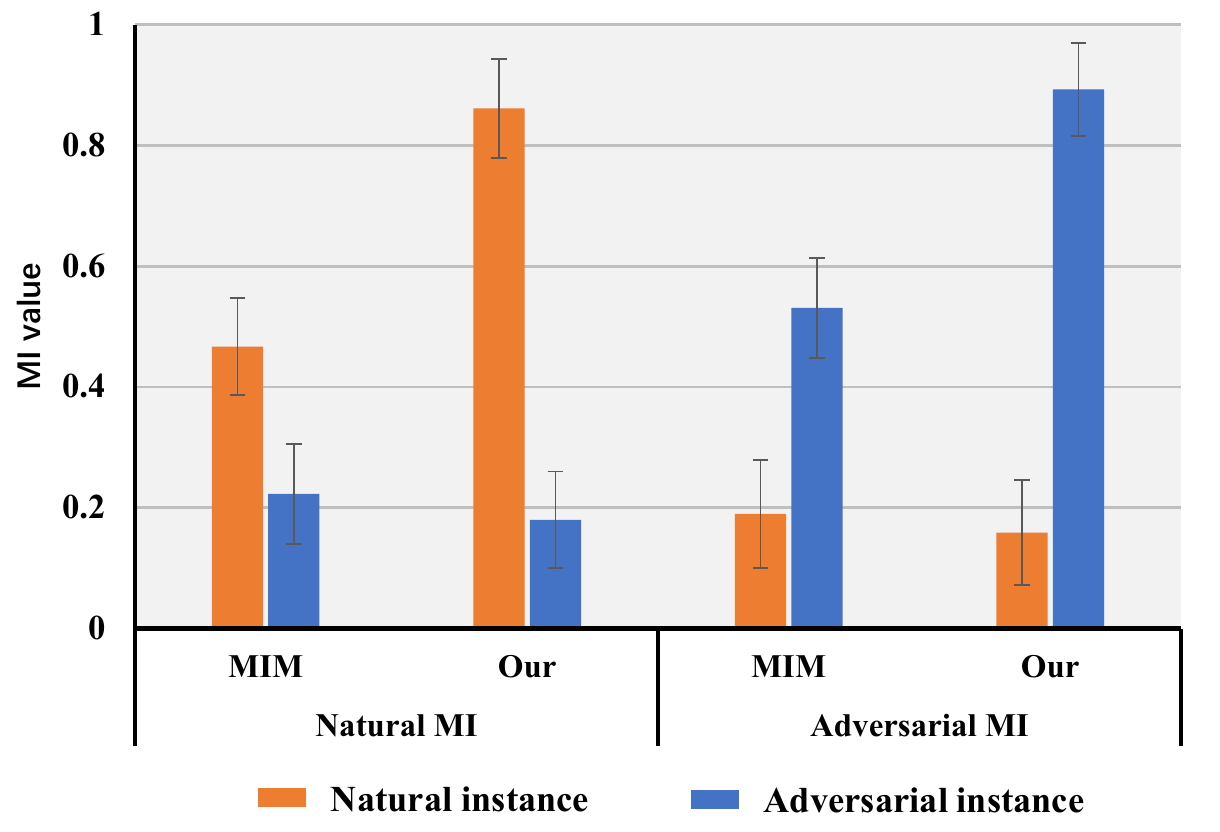}}
\caption{The performances of MI estimation networks trained by \cref{eq4} (MIM) and \cref{eq5} (Our). The left half is the estimated natural MI, and the right half is the adversarial MI.}
\label{fig4}
\vskip -0.35in
\end{center}
\end{figure}

\begin{table*}[hbtp]
\caption{Adversarial accuracy (percentage) of defense methods against white-box attacks on \textit{CIFAR-10} and \textit{Tiny-ImageNet}. The target model is ResNet-18.}
\label{tab1}
\renewcommand\tabcolsep{6pt}
\renewcommand\arraystretch{1.1}
\begin{center}
\begin{small}
\begin{tabular}{l|l|ccccc|cccc}
\hline
\multirow{2}{*}{Dataset} &\multirow{2}{*}{Defense}  &\multicolumn{5}{c|}{$L_{\infty}$-norm} &\multicolumn{4}{c}{$L_{2}$-norm} \\
& & None & PGD-40 & AA & FWA-40 &TI-DIM & None &PGD-40& CW & DDN \\ \hline
\multirow{7}{*}{CIFAR-10} &Standard & 83.39 & 42.38 & 39.01 & 15.44 &55.63 &83.97 &61.69 & 30.96 &29.34 \\
&WMIM & 80.32 & 40.76 & 36.05 & 12.14 &53.10 &81.29 &58.36 & 28.41 &27.13 \\
&NAMID & \textbf{83.41} & \textbf{44.79} & \textbf{39.26} & \textbf{15.67} &\textbf{58.23} &\textbf{84.35} &\textbf{62.38} & \textbf{34.48} &\textbf{32.41}\\ \cdashline{2-11}[3pt/5pt]
&TRADES & \textbf{80.70} & 46.29 & 42.71 & 20.54 &57.06 &83.72 &63.17 & 33.81 &32.06 \\
&NAMID\_T &80.67  &\textbf{47.53}  &\textbf{43.39}  &\textbf{21.17} &\textbf{59.13} &\textbf{84.19} &\textbf{64.75} &\textbf{35.41} &\textbf{34.27} \\ \cdashline{2-11}[3pt/5pt]
&MART & 78.21 & 50.23 & 43.96 & 25.56 &58.62  &83.36 &65.38 & 35.57 &34.69  \\
&NAMID\_M &\textbf{78.38} &\textbf{51.69}  &\textbf{44.42}  &\textbf{25.64} &\textbf{61.26} &\textbf{84.07} &\textbf{66.03} &\textbf{36.19} &\textbf{35.76} \\ \hline
\multirow{7}{*}{Tiny-ImageNet} &Standard & 48.40 & 17.35 & 11.27 & 10.29 &27.84 &49.57 &26.19 &12.73 &11.25 \\
&WMIM & 47.43 & 16.50 & 9.87 & 9.25 & 25.19 &48.16 &24.10 &11.35 &10.16  \\
&NAMID & \textbf{48.41} & \textbf{18.67} & \textbf{12.29} & \textbf{11.32} &\textbf{29.37}  &\textbf{49.65} &\textbf{28.13} & \textbf{14.29} &\textbf{12.57} \\ \cdashline{2-11}[3pt/5pt]
&TRADES & \textbf{48.25} & 19.17 & 12.36 & 10.67 &29.64  & 48.83 & 27.16 & 13.28 &12.34 \\
&NAMID\_T &48.21  &\textbf{20.12}  &\textbf{12.86}  &\textbf{14.91} &\textbf{30.81}  &\textbf{49.07} &\textbf{28.83} &\textbf{14.47} &\textbf{13.91} \\ \cdashline{2-11}[3pt/5pt]
&MART & \textbf{47.83} & 20.90 & 15.57 & 12.95 &30.71 &48.56 &27.98 & 14.36 &13.79  \\
&NAMID\_M &47.80  &\textbf{21.23}  &\textbf{15.83}  &\textbf{15.09} & \textbf{31.59} &\textbf{48.72} &\textbf{29.14} &\textbf{15.06} &\textbf{14.23} \\ \hline
\end{tabular}
\end{small}
\end{center}
\vskip -0.15in
\end{table*}

We use the test data from \textit{CIFAR-10} to evaluate the performance. For the natural MI, we offset the estimated MI so that the worst-case of the natural MI equals 0, and calculate the average of all samples. Similarly, we offset the estimated MI so that the worst-case of the adversarial MI equals 0. Note that for a fair comparison, we use selected samples to train the estimation networks for both methods. As shown in \cref{fig4}, the results demonstrate that the optimization mechanism could help adequately represent the inherent difference in the natural MI and the adversarial MI between the natural sample and the adversarial sample.

\subsection{Robustness evaluation and analysis}
\label{section4.3}

To demonstrate the effectiveness of our adversarial defense algorithm, we evaluate the adversarial accuracy using white-box and black-box adversarial attacks, respectively.

\noindent\textbf{White-box attacks.} 
In the white-box setting, all attacks can access the architectures and parameters of target models. We evaluate the robustness by exploiting six types of adversarial attacks for both \textit{CIFAR-10} and \textit{Tiny-ImageNet}: $L_{\infty}$-norm PGD, FWA, AA, TI-DIM attacks and $L_{2}$-norm PGD, DDN, CW attacks. The average natural accuracy (i.e., the results in the third column) and the average adversarial accuracy of defenses are shown in \cref{tab1}.

The results show that our method (i.e., NAMID) can achieve better robustness compared with \textit{Standard AT}. The performance of our method on the natural accuracy is competitive (83.39\% vs. 83.41\%), and it provides more gains on adversarial accuracy (e.g., 5.69\% against PGD-40). Compared with \textit{WMIM}, the results show that our proposed strategy of disentangling the standard MI into the natural MI and the adversarial MI is effective. The standard deviation is shown in \cref{appendix_5_2}.

Note that the default adversarial training loss in our method (i.e., $\mathcal{L}_{adv}$ in \cref{eq9}) is the same as \textit{Standard AT}. To avoid the bias caused by different adversarial training methods, we apply the adversarial training losses of \textit{TRADES} and \textit{MART} to our method respectively (i.e., NAMID\_T and NAMID\_M). As shown in \cref{tab1}, the results show that our method can improve the adversarial accuracy (e.g., the accuracy against PGD is improved by 2.68\% and 2.91\% compared with \textit{TRADES} and \textit{MART} on \textit{CIFAR-10}).

\noindent\textbf{Black-box attacks.}
Block-box adversarial instances are crafted by attacking a surrogate model. We use a VggNet-19 \cite{he2016deep} as the surrogate model. The surrogate models and the defense models are trained separately. We use \textit{Standard AT} method to train the surrogate model and use PGD, AA and FWA to generate adversarial test data. The performances of our defense method is reported in \cref{tab2}. The results show that our method is a practical strategy for real scenarios, which can protect the target model from black-box attacks by malicious adversaries.

\begin{table}[t]
\caption{Adversarial accuracy (percentage) of defense methods against black-box attacks on \textit{CIFAR-10}. The target model is ResNet-18 and the surrogate model is adversarially trained VggNet-19. We show the most successful defense with \textbf{bold}.}
\label{tab2}
\renewcommand\tabcolsep{8pt}
\renewcommand\arraystretch{1.05}
\begin{center}
\begin{small}
\begin{tabular}{l|cccc}
\hline
Defense  & None & PGD-40 & AA & FWA-40 \\ \hline
Standard & 83.39    & 65.88    & 60.93   & 56.42       \\
WMIM     & 80.32     & 62.79       & 57.86   &  53.05      \\
NAMID      & \textbf{83.41}     & \textbf{69.57}       & \textbf{63.72}   & \textbf{59.30}       \\ \hline
\end{tabular}
\end{small}
\end{center}
\vskip -0.25in
\end{table}

%Note that the robustness improvement of DMIAD is not caused by \textit{obfuscated gradients} \cite{}. Three phenomenons can demonstrate this: (i) stronger attacks (e.g., AA) have higher attack success rates (lower classification accuracies) than weaker test attacks (e.g., PGD); (ii) white-box attacks have higher attack success rates than back-box attacks (see \cref{tab2} with \cref{tab1}); (iii) the gradient-free attack SPSA \cite{} does not cause higher attack success rates than gradient-based attacks (e.g.,PGD). The SPSA attack is an additional check and it results confirms that the robustness of our method are not due to gradient masking.

\subsection{Ablation study}
\label{section4.4}
\begin{figure}[t]
\begin{center}
\vskip 0.1in
\centerline{\includegraphics[width=2.5 in]{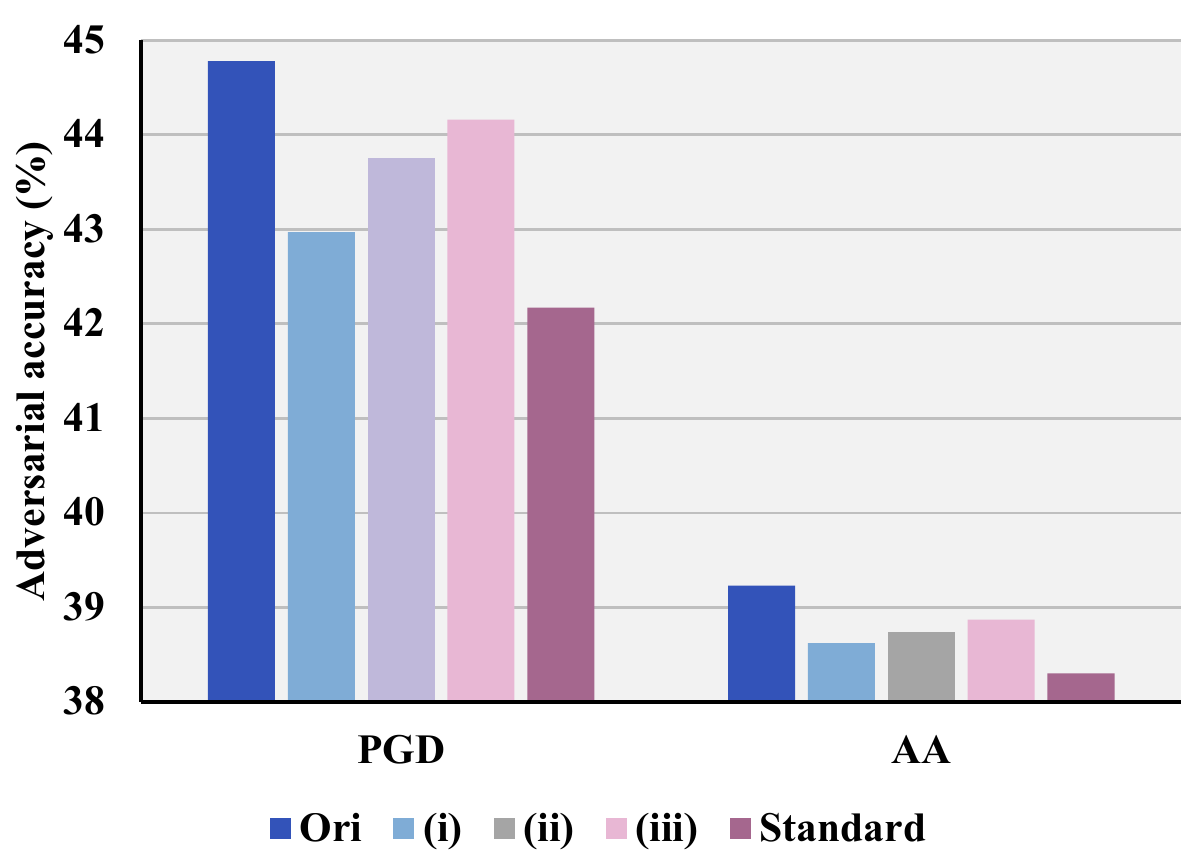}}
\caption{The ablation study.The bars with different colors represent the performance under different settings. Among them, 'Ori' denotes our method NAMID and 'Standard' denotes \textit{Standard AT}.}
\label{fig5}
\vskip -0.35in
\end{center}
\end{figure}

To clearly elucidate the role of each component of our method in improving adversarial robustness, we conduct ablation studies in three different settings: (i) removing the adversarial MI; (ii) removing the natural MI; and (iii) setting the hyperparameter $\lambda$ (in \cref{eq8}) to 0. We use $L_{\infty}$-norm PGD and FWA attacks to evaluate the performances of these variants. As shown in \cref{fig5}, the results demonstrate that each component of our method contributes positively to improving adversarial accuracy.

\section{Conclusion}
\label{section5}
To the best of our knowledge, the dependence between the output of the target model and input adversarial samples have not been well studied. In this paper, we investigate the dependence from the perspective of information theory. Considering that adversarial samples contain natural and adversarial patterns, we propose to disentangle the standard MI into the natural MI and the adversarial MI to explicitly measure the dependence of the output on the different patterns. We design a neural network-based method to train two MI estimation networks to estimate the natural MI and the adversarial MI. Based on the above MI estimation, we develop an adversarial defense algorithm called natural-adversarial mutual information-based defense (NAMID) to enhance the adversarial robustness. The empirical results demonstrate that our defense method can provide effective protection against multiple adversarial attacks. Our work provides a new adversarial defense strategy for the community of adversarial learning. In future, we will design more efficient mechanisms for training MI estimators and further optimize the natural-adversarial MI-based defense to improve the performance against stronger attacks. In addition, 

\section{Acknowledgements}
This work was supported in part by the National Key Research and Development Program of China under Grant 2018AAA0103202, in part by the National Natural Science Foundation of China under Grant 61922066, 61876142, 62036007, 62006202, 61922066, 61876142, 62036007, and 62002090, in part by the Technology Innovation Leading Program of Shaanxi under Grant 2022QFY01-15, in part by Open Research Projects of Zhejiang Lab under Grant 2021KG0AB01, in part by the RGC Early Career Scheme No. 22200720, in part by Guangdong Basic and Applied Basic Research Foundation No. 2022A1515011652, in part by Australian Research Council Projects DE-190101473, IC-190100031, and DP-220102121, in part by the Fundamental Research Funds for the Central Universities, and in part by the Innovation Fund of Xidian University. The authors thank the reviewers and the meta-reviewer for their helpful and constructive comments on this work.

\bibliography{example_paper}
\bibliographystyle{icml2022}

%%%%%%%%%%%%%%%%%%%%%%%%%%%%%%%%%%%%%%%%%%%%%%%%%%%%%%%%%%%%%%%%%%%%%%%%%%%%%%%
%%%%%%%%%%%%%%%%%%%%%%%%%%%%%%%%%%%%%%%%%%%%%%%%%%%%%%%%%%%%%%%%%%%%%%%%%%%%%%%
% APPENDIX
%%%%%%%%%%%%%%%%%%%%%%%%%%%%%%%%%%%%%%%%%%%%%%%%%%%%%%%%%%%%%%%%%%%%%%%%%%%%%%%
%%%%%%%%%%%%%%%%%%%%%%%%%%%%%%%%%%%%%%%%%%%%%%%%%%%%%%%%%%%%%%%%%%%%%%%%%%%%%%%
\newpage
\appendix
\onecolumn
\section{Related work on mutual information}
\label{appendix_1}
Mutual information (MI) is an entropy-based measure that can reflect the dependence degree between variables. The definition of MI is as follows.

\noindent \textbf{Definition 1.} 
Let $(X,Z)$ be a pair of random variables and $\mathcal{X} \times \mathcal{Z}$ be their feature space. The MI of $(X,Z)$ is defined as:

\begin{equation}
\label{eq10}
I(X;Z)=\int_{Z} \int_{\mathcal{X}} p_{X Z}(x, z) \log \left(\frac{p_{X Z}(x, z)}{p_{X}(x) p_{Z}(z)}\right) d x d z \text{,}
\end{equation}

where $p_{X Z}$ is the joint probability density function of $(X, Z)$, and $p_X$, $p_Z$ are the marginal probability density functions of $X$ and $Z$ respectively.

Intuitively, $I(X;Z)$ could reflect how well one can predict $Z$ from $X$ (and $X$ from $Z$, since it is symmetrical \cite{zhu2020learning}). By definition, $I(X;Z) = 0$ if $X$ and $Z$ are independent; A larger MI typically indicates a stronger dependence between the two variables.

Various methods have been proposed for estimating MI \cite{moon1995estimation,darbellay1999estimation,kandasamy2015nonparametric,moon2017ensemble}. A representative and efficient estimator is the mutual information neural estimator (MINE) \cite{belghazi2018mutual}:

\begin{equation}
\label{eq11}
\widehat{I}_{\omega}(X ; Z)=\sup _{\omega \in \Omega} \mathbb{E}_{\widehat{\mu}_{X Z}^{(n)}}\left[T_{\omega}\right]-\log \left(\mathbb{E}_{\widehat{\mu}_{X}^{(n)} \otimes \widehat{\mu}_{Z}^{(n)}}\left[\exp \left(T_{\omega}\right)\right]\right) \text{,}
\end{equation}
where $T_{\omega}: \mathcal{X} \times \mathcal{Z} \rightarrow \mathbb{R}$ is a function modeled by a neural network with parameters $\omega \in \Omega$. $\widehat{\mu}_{X Z}^{(n)}$ , $\mu_{X}^{(n)}$ and $\mu_{Z}^{(n)}$ are the empirical distributions of random variables $(X,Z)$, $X$ and $Z$ respectively, associated to n $i.i.d.$ samples. MINE is empirically demonstrated its superiority in estimation accuracy and efficiency, and proved that it is strongly consistent with the true MI. 

Besides, the work in \citet{hjelm2018learning} points that using the complete input to estimate MI is often insufficient for classification task. Instead, estimating MI between the high-level feature and local patches of the input is more suitable. This work propose a local DIM estimator to estimate MI:

\begin{equation}
\label{eq12}
\widehat{I}_{\psi,\omega}(X ; Z)= \widehat{I}_{\omega}(C_{\psi}(X) ; Z) \text{,}
\end{equation}
where $C_{\psi}$ is a encoder modeled by a neural network with parameter $\psi$. It encode the input to a local feature map that reflects useful structure in the data (e.g., spatial locality). In this paper, we refer the local DIM estimator to estimate the natural MI and the adversarial MI. 

\section{Proof-of-concept experiment}
\label{appendix_2}

We randomly select 1,000 natural test samples from \textit{CIFAR-10} as the experimental data. We exploit three representative adversarial attacks (i.e., PGD \cite{madry2017towards}, AA \cite{croce2020reliable} and FWA \cite{wu2020stronger}) with different iteration numbers to generative adversarial samples respectively. We use a naturally trained ResNet-18 network \cite{he2016deep} as the target model $h$. We respectively estimate the standard MI of the natural sample $\widehat{I}(x, h(x))$, the standard MI of the adversarial sample $\widehat{I}(\tilde{x}, h(\tilde{x}))$ and the MI between the natural sample and the output for the corresponding adversarial sample $\widehat{I}(x, h(\tilde{x}))$ via the local DIM estimator. Note that the natural MI of the natural sample is equivalent to its standard MI. We offset the estimated MI so that the worst-case of the estimated MI equals 0, and then calculate the average of all samples for $\widehat{I}(x, h(x))$, $\widehat{I}(\tilde{x}, h(\tilde{x}))$ and $\widehat{I}(x, h(\tilde{x}))$ respectively.

\section{Proof of Theorem 1}
\label{appendix_3}
In this section, we provide the proof of the theoretical result in \cref{section3.2.1}.

\noindent \textbf{Theorem 1.} 
Let $X, \widetilde{X}, N, Z$ denote four random variables respectively, where $\widetilde{X}=X+N$. Let $\widetilde{\mathcal{X}}$ be the feature space of $\widetilde{X}$ and $\mathcal{Z}$ be the feature space of $Z$. 
%$\widetilde{X}$ and $Z$ are random variables that follow the marginal distribution $\mu_{\widetilde{X}}$ and $\mu_{Z}$ respectively. 
Then, for any function $h: \widetilde{\mathcal{X}} \rightarrow \mathcal{Z}$, we have 
\begin{equation}
\label{eq13}
I(\widetilde{X};Z) = I(X;Z) + I(N;Z) - I(X;N;Z) + H(Z|X,N) - H(Z|\widetilde{X})  \text{,}
\end{equation}
where $I(\cdot;\cdot)$ denotes the mutual information between two variables and $I(\cdot;\cdot;\cdot)$ denotes the mutual information between three variables.

\noindent \textbf{Proof.}
According to the relationship between information entropy and mutual information, we first provide the following \textbf{Lemma 1}:

\begin{equation}
\label{eq14}
\left\{\begin{array}{l}
I(\widetilde{X};Z)=H(Z)-H(Z|\widetilde{X}) \\
I(X;Z)=H(Z)-H(Z|X) \\
I(N;Z)=H(Z)-H(Z|N) 
\end{array}\right. \text{,}
\end{equation}

where $H(\cdot)$ denotes the information entropy, and $H(\cdot|\cdot)$ denotes the conditional information entropy.

According to Lemma 1, we have:
\begin{equation}
\label{eq15}
\begin{aligned}
I(X;Z)+I(N;Z)&=H(Z)-H(Z|X)+H(Z)-H(Z|N) \\
             &=2 \, H(Z) - [H(Z|X) + H(Z|N)] \text{.}
\end{aligned}
\end{equation}

According to the theorem of conditional mutual information in probability theory, we have:

\begin{equation}
\label{eq16}
\begin{aligned}
H(Z|X) + H(Z|N) &= [H(Z|X,N) + I(Z;N|X)] + [H(Z|X,N) + I(X;Z|N)] \\
                &= 2 \, H(Z|X,N) + I(Z;N|X) + I(X;Z|N) \\
                &=[H(Z|X,N) + I(Z;N|X) + I(X;Z|N) + I(X;N;Z)] \\
                & \; \; \; \; \; + H(Z|X,N) - I(X;N;Z) \\
                &= H(Z) + H(Z|X,N) - I(X;N;Z) \text{.}
\end{aligned}
\end{equation}

Combining \cref{eq15} and \cref{eq16}, we have: 
\begin{equation}
\label{eq17}
\begin{aligned}
I(X;Z)+I(N;Z) &= 2 \, H(Z) - [H(Z) + H(Z|X,N) - I(X;N;Z)] \\
              &= H(Z) - H(Z|X,N) + I(X;N;Z) \\
              &= I(\widetilde{X};Z) + H(Z|\widetilde{X}) - H(Z|X,N) + I(X;N;Z) \text{.}
\end{aligned}
\end{equation}

Finally, we have:
\begin{equation}
\label{eq18}
\begin{aligned}
 I(\widetilde{X};Z) = I(X;Z) + I(N;Z) - I(X;N;Z) + H(Z|X,N) - H(Z|\widetilde{X}) \text{,}
\end{aligned}
\end{equation}
which completes the proof.

\section{The assumptions in Eq. (3).}
\label{appendix_4}
The adversarial noise usually depends on the natural instance, that is, there is also a dependence between the adversarial noise and the natural sample. The MI between the natural sample and the adversarial noise  (i.e., $I(X;N)$) may cause some effects on the disentanglement of standard MI in \cref{eq3}. In fact, the effect of this MI is tiny. We illustrate this using two Venn diagrams.

\begin{figure}[hbtp]
\begin{center}
\vskip 0.1in
\centerline{\includegraphics[width=2.5in]{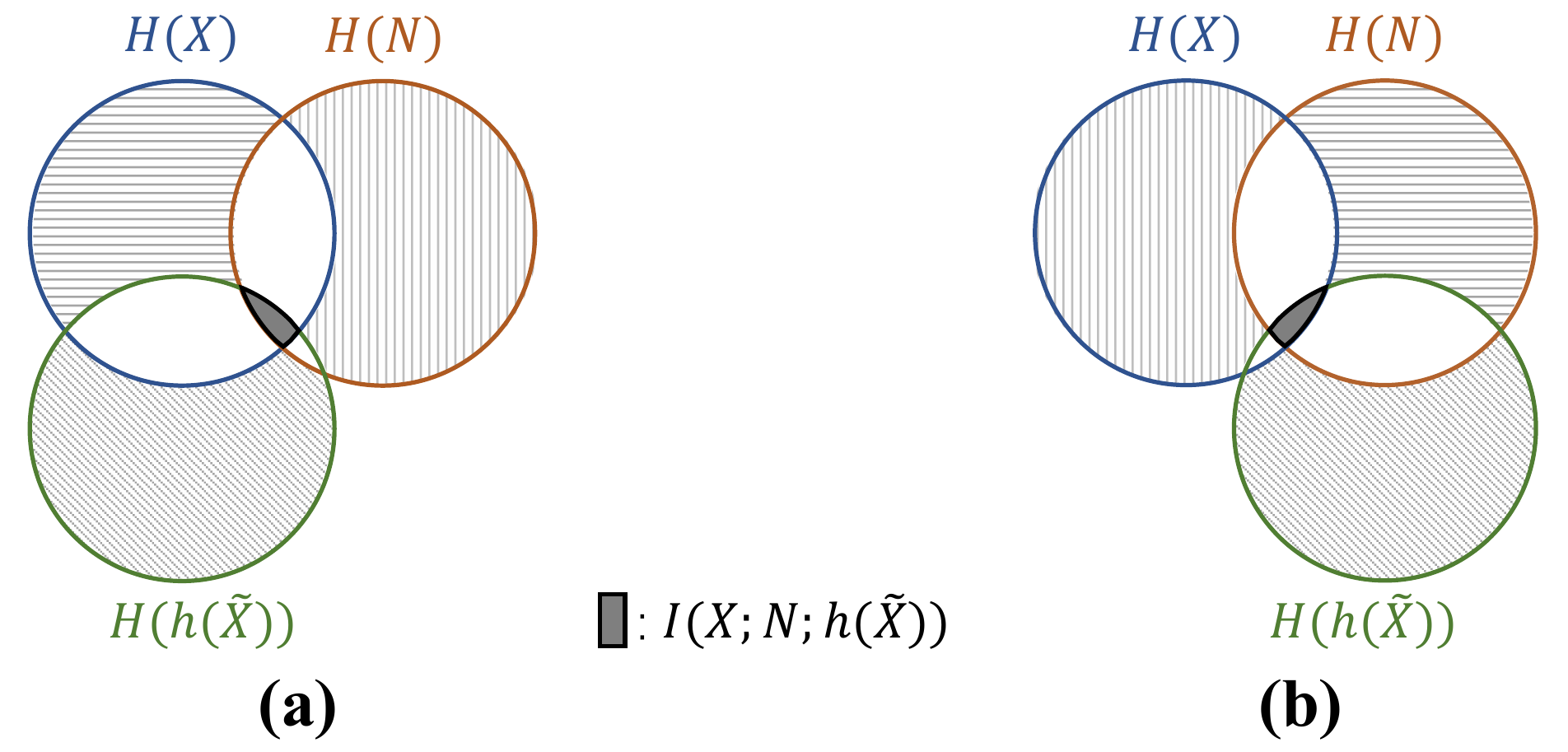}}
\caption{Venn diagrams for mutual information.}
\label{fig6}
\vskip -0.3in
\end{center}
\end{figure}

As shown in \cref{fig6}, since the adversarial noise is usually crafted based on the natural instance, $H(X)$ and $H(N)$ have a large overlap ($H$ denotes entropy). Considering that predictions are zero-sum (\textit{i.e.}, either right or wrong), and the effects of natural and adversarial patterns on outputs are mutually exclusive, $H(h(\widetilde{X}))$ either has a large overlap with $H(X)$ and a small overlap with $H(N)$ (\cref{fig6}(a)), or has a large overlap with $H(N)$ and a small overlap with $H(X)$ (\cref{fig6}(b)). In both cases, the overlap common to $H(X)$, $H(N)$, and $H(h(\widetilde{X}))$ (\textit{i.e.}, $I(X;N;h(\widetilde{X}))$) is small. Thus, $I(X;N)$ does not significantly affect \cref{eq3}.

In addition, for the difference between $H(Z|X,N)$ and $H(Z|\widetilde{X})$, if we can recover the natural sample $X$ and the adversarial noise $N$ from $X+N$, then $X+N$ can contain all the information of $X$ and $N$, so the difference $H(Z|X,N)-H(Z|X+N)$ equals 0. Of course, this recovery process is difficult to precisely implement. Considering that we use $\widetilde{X}=X+N$ to generate adversarial sample in our work, we can precisely obtain $X$ and $N$, so we assume that the difference $H(Z|X,N)-H(Z|X+N)$ may tend to be small.

\section{Experimental results}
\label{appendix_5}
\subsection{Different MI estimation networks}
\label{appendix_5_1}
\begin{figure}[hbtp]
\begin{center}
\vskip 0.1in
\centerline{\includegraphics[width=2.5in]{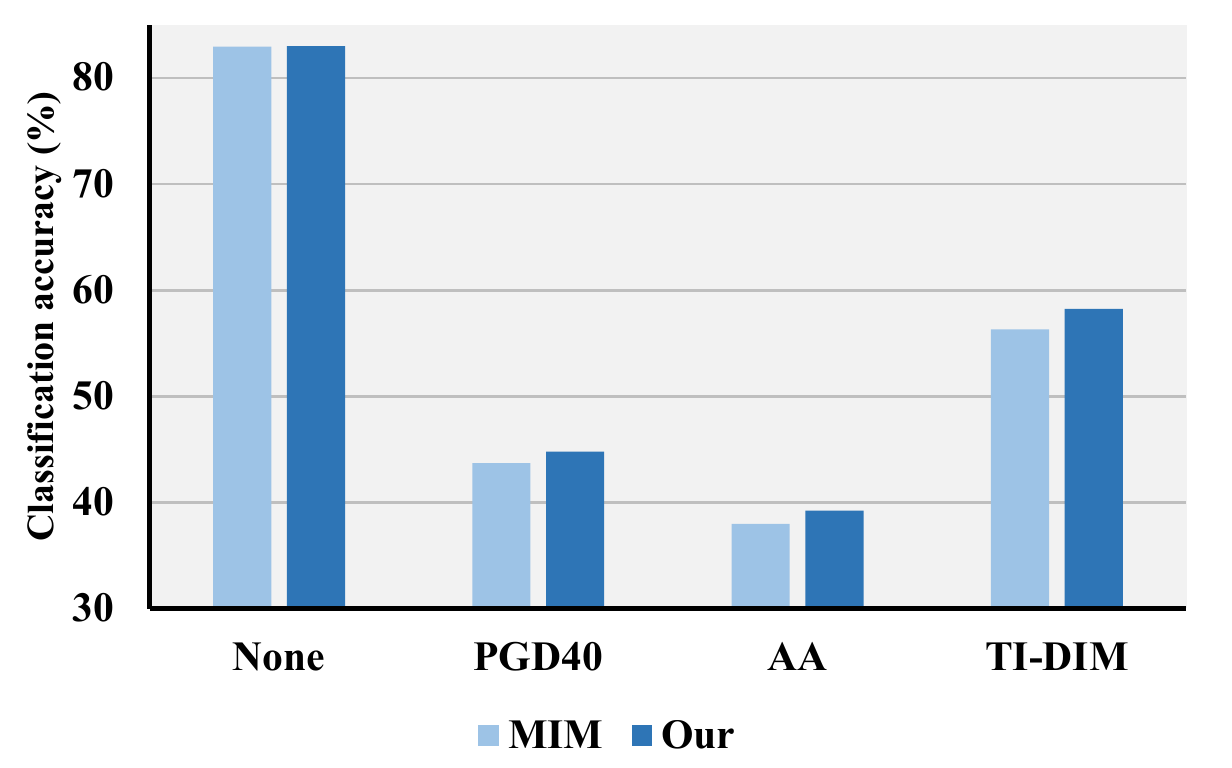}}
\caption{The performances of the defense based on different estimation networks. 'MIM' denotes the defense model based on the MI estimation network trained by \cref{eq4}. 'Our' denotes the defense model based on the MI estimation network trained by \cref{eq5}.}
\label{fig7}
\vskip -0.3in
\end{center}
\end{figure}

We use the MI estimation network trained by \cref{eq4} and \cref{eq5} to train our adversarial defense model, respectively. The results are shown in \cref{fig7}. From the results, it can be seen that our proposed method for training the MI estimation network can better facilitate the defense model to improve the adversarial accuracy.

\subsection{Defense against adversarial attacks}
\label{appendix_5_2}
We show the performances of our defense algorithm against white-box $L_{\infty}$-norm attacks and $L_2$-norm attacks in \cref{tab3} and \cref{tab4}, respectively. The performances of our defense algorithm against black-box $L_{\infty}$-norm attacks and $L_2$-norm attacks are shown in \cref{tab5} and \cref{tab6}.

\begin{table*}[hbtp]
\caption{Adversarial accuracy (percentage) of defense methods against $L_\infty$-norm white-box attacks on \textit{CIFAR-10} and \textit{Tiny-ImageNet}. The target model is ResNet-18. The adversarial training data is crafted by using $L_{\infty}$-norm PGD-10.}
\label{tab3}
\renewcommand\tabcolsep{8pt}
\renewcommand\arraystretch{1.1}
\begin{center}
\begin{small}
\begin{tabular}{l|l|ccccc}
\hline
Dataset &Defense & None & PGD-40 & AA & FWA-40 &TI-DIM  \\ \hline
\multirow{7}{*}{CIFAR-10} &Standard & 83.39$\pm$0.95 & 42.38$\pm$0.56 & 39.01$\pm$0.51 & 15.44$\pm$0.32 &55.63$\pm$0.67 \\
&WMIM & 80.32$\pm$0.60 & 40.76$\pm$0.45 & 36.05$\pm$0.57 & 12.14$\pm$0.46 &53.10$\pm$0.53  \\
&NAMID & \textbf{83.41$\pm$0.56} & \textbf{44.79$\pm$0.50} & \textbf{39.26$\pm$0.56} & \textbf{15.67$\pm$0.47} &\textbf{58.23$\pm$0.63} \\ \cdashline{2-7}[3pt/5pt]
&TRADES & \textbf{80.70$\pm$0.63} & 46.29$\pm$0.59 & 42.71$\pm$0.49 & 20.54$\pm$0.47 &57.06$\pm$0.51 \\
&NAMID\_T &80.67$\pm$0.73  &\textbf{47.53$\pm$0.59}  &\textbf{43.39$\pm$0.83}  &\textbf{21.17$\pm$0.44} &\textbf{59.13$\pm$0.72} \\ \cdashline{2-7}[3pt/5pt]
&MART & 78.21$\pm$0.65 & 50.23$\pm$0.70 & 43.96$\pm$0.67 & 25.56$\pm$0.61 &58.62$\pm$0.65 \\
&NAMID\_M &\textbf{78.38$\pm$0.68} &\textbf{51.69$\pm$0.51}  &\textbf{44.42$\pm$0.71}  &\textbf{25.64$\pm$0.49} &\textbf{61.26$\pm$0.67} \\ \hline
\multirow{7}{*}{Tiny-ImageNet} &Standard & 48.40$\pm$0.68 & 17.35$\pm$0.56 & 11.27$\pm$0.53 & 10.29$\pm$0.47 &27.84$\pm$0.46 \\
&WMIM & 47.43$\pm$0.56 & 16.50$\pm$0.72 & 9.87$\pm$0.57 & 9.25$\pm$0.45 & 25.19$\pm$0.47 \\
&NAMID & \textbf{48.41$\pm$0.47} & \textbf{18.67$\pm$0.59} & \textbf{12.29$\pm$0.43} & \textbf{11.32$\pm$0.52} &\textbf{29.37$\pm$0.67} \\ \cdashline{2-7}[3pt/5pt]
&TRADES & \textbf{48.25$\pm$0.71} & 19.17$\pm$0.58 & 12.63$\pm$0.51 & 10.67$\pm$0.68 &29.64$\pm$0.43 \\
&NAMID\_T &48.21$\pm$0.54  &\textbf{20.12$\pm$0.66}  &\textbf{12.86$\pm$0.57}  &\textbf{14.91$\pm$0.41} &\textbf{30.81$\pm$}0.60 \\ \cdashline{2-7}[3pt/5pt]
&MART & \textbf{47.83$\pm$0.65} & 20.90$\pm$0.59 & 15.57$\pm$0.52 & 12.95$\pm$0.49 &30.71$\pm$0.41 \\
&NAMID\_M &47.80$\pm$0.52  &\textbf{21.23$\pm$0.49}  &\textbf{15.83$\pm$0.50}  &\textbf{15.09$\pm$0.57} & \textbf{31.59$\pm$0.63} \\ \hline
\end{tabular}
\end{small}
\end{center}
\vskip -0.1in
\end{table*}

\begin{table*}[hbtp]
\caption{Adversarial accuracy (percentage) of defense methods against $L_2$-norm white-box attacks on \textit{CIFAR-10} and \textit{Tiny-ImageNet}. The target model is ResNet-18. The adversarial training data is crafted by using $L_2$-norm PGD-10.}
\label{tab4}
\renewcommand\tabcolsep{8pt}
\renewcommand\arraystretch{1.1}
\begin{center}
\begin{small}
\begin{tabular}{l|l|cccc}
\hline
Dataset &Defense & None &PGD-40& CW & DDN \\ \hline
\multirow{7}{*}{CIFAR-10} &Standard &83.97$\pm$0.51 &61.69$\pm$0.74 & 30.96$\pm$0.53 &29.34$\pm$0.70 \\
&WMIM &81.29$\pm$0.73 &58.36$\pm$0.51 & 28.41$\pm$0.47 &27.13$\pm$0.54 \\
&NAMID &\textbf{84.35$\pm$0.76} &\textbf{62.38$\pm$0.70} & \textbf{34.48$\pm$0.58} &\textbf{32.41$\pm$0.69}\\ \cdashline{2-6}[3pt/5pt]
&TRADES &83.72$\pm$0.66 &63.17$\pm$0.43 & 33.81$\pm$0.71 &32.06$\pm$0.75 \\
&NAMID\_T &\textbf{84.19$\pm$0.56} &\textbf{64.75$\pm$0.84} &\textbf{35.41$\pm$0.65} &\textbf{34.27$\pm$0.75} \\ \cdashline{2-6}[3pt/5pt]
&MART &83.36$\pm$0.47 &65.38$\pm$0.49 & 35.57$\pm$0.66 &34.69$\pm$0.62  \\
&NAMID\_M &\textbf{84.07$\pm$0.65} &\textbf{66.03$\pm$0.53} &\textbf{36.19$\pm$0.46} &\textbf{35.76$\pm$0.69} \\ \hline
\multirow{7}{*}{Tiny-ImageNet} &Standard &49.57$\pm$0.43 &26.19$\pm$0.46 &12.73$\pm$0.59 &11.25$\pm$0.42 \\
&WMIM &48.16$\pm$0.67 &24.10$\pm$0.43 &11.35$\pm$0.61 &10.16$\pm$0.74  \\
&NAMID &\textbf{49.65$\pm$0.40} &\textbf{28.13$\pm$0.69} & \textbf{14.29$\pm$0.48} &\textbf{12.57$\pm$0.70} \\ \cdashline{2-6}[3pt/5pt]
&TRADES & 48.83$\pm$0.63 & 27.16$\pm$0.43 & 13.28$\pm$0.37 &12.34$\pm$0.52 \\
&NAMID\_T &\textbf{49.07$\pm$0.40} &\textbf{28.83$\pm$0.62} &\textbf{14.47$\pm$0.56} &\textbf{13.91$\pm$0.70} \\ \cdashline{2-6}[3pt/5pt]
&MART &48.56$\pm$0.53 &27.98$\pm$0.55 & 14.36$\pm$0.64 &13.79$\pm$0.51  \\
&NAMID\_M &\textbf{48.72$\pm$0.75} &\textbf{29.14$\pm$0.43} &\textbf{15.06$\pm$0.52} &\textbf{14.23$\pm$0.39} \\ \hline
\end{tabular}
\end{small}
\end{center}
\vskip -0.1in
\end{table*}

\begin{table*}[t]
\caption{Adversarial accuracy (percentage) of defense methods against $L_\infty$-norm black-box attacks on \textit{CIFAR-10} and \textit{Tiny-ImageNet}. The target model is ResNet-18. The surrogate model is VggNet-19. The adversarial training data is crafted by using $L_{\infty}$-norm PGD-10.}
\label{tab5}
\renewcommand\tabcolsep{8pt}
\renewcommand\arraystretch{1.1}
\begin{center}
\begin{small}
\begin{tabular}{l|l|ccccc}
\hline
Dataset &Defense & None & PGD-40 & AA & FWA-40 &TI-DIM  \\ \hline
\multirow{3}{*}{CIFAR-10} &Standard & 83.39$\pm$0.95 & 65.88$\pm$0.47 & 60.93$\pm$0.49 & 56.42$\pm$0.50 &70.17$\pm$0.73 \\
&WMIM & 80.32$\pm$0.60 & 62.79$\pm$0.63 & 57.86$\pm$0.56 & 53.05$\pm$0.86 &67.41$\pm$0.84  \\
&NAMID & \textbf{83.41$\pm$0.56} & \textbf{69.57$\pm$0.72} & \textbf{63.72$\pm$0.71} & \textbf{59.30$\pm$0.63} &\textbf{72.83$\pm$0.81} \\ \hline
\multirow{3}{*}{Tiny-ImageNet} &Standard & 48.40$\pm$0.68 & 36.16$\pm$0.37 & 32.50$\pm$0.46 & 30.47$\pm$0.32 &39.63$\pm$0.73 \\
&WMIM & 47.43$\pm$0.56 & 33.09$\pm$0.69 & 30.87$\pm$0.70 & 28.49$\pm$0.38 & 35.34$\pm$0.62 \\
&NAMID & \textbf{48.41$\pm$0.47} & \textbf{38.10$\pm$0.62} & \textbf{34.57$\pm$0.54} & \textbf{31.86$\pm$0.53} &\textbf{39.75$\pm$0.72} \\ \hline
\end{tabular}
\end{small}
\end{center}
\vskip -0.1in
\end{table*}

\begin{table*}[t]
\caption{Adversarial accuracy (percentage) of defense methods against $L_2$-norm black-box attacks on \textit{CIFAR-10} and \textit{Tiny-ImageNet}. The target model is ResNet-18. The surrogate model is VggNet-19. The adversarial training data is crafted by using $L_2$-norm PGD-10.}
\label{tab6}
\renewcommand\tabcolsep{8pt}
\renewcommand\arraystretch{1.1}
\begin{center}
\begin{small}
\begin{tabular}{l|l|cccc}
\hline
Dataset &Defense & None &PGD-40& CW & DDN \\ \hline
\multirow{3}{*}{CIFAR-10} &Standard &83.97$\pm$0.51 &75.69$\pm$0.85 & 68.53$\pm$0.72 &65.19$\pm$0.87 \\
&WMIM &81.29$\pm$0.73 &72.91$\pm$0.66 & 64.70$\pm$0.71 &62.16$\pm$0.76 \\
&NAMID &\textbf{84.35$\pm$0.76} &\textbf{78.43$\pm$0.77} & \textbf{70.25$\pm$0.61} &\textbf{68.37$\pm$0.54}\\ \hline
\multirow{3}{*}{Tiny-ImageNet} &Standard &49.57$\pm$0.43 &40.36$\pm$0.65 &34.13$\pm$0.57 &32.79$\pm$0.39 \\
&WMIM &48.16$\pm$0.67 &39.56$\pm$0.43 &31.79$\pm$0.69 &30.25$\pm$ 0.35 \\
&NAMID &\textbf{49.65$\pm$0.40} &\textbf{41.78$\pm$0.48} & \textbf{36.43$\pm$0.46} &\textbf{34.61$\pm$0.49} \\ \hline
\end{tabular}
\end{small}
\end{center}
\vskip -0.1in
\end{table*}

%%%%%%%%%%%%%%%%%%%%%%%%%%%%%%%%%%%%%%%%%%%%%%%%%%%%%%%%%%%%%%%%%%%%%%%%%%%%%%%
%%%%%%%%%%%%%%%%%%%%%%%%%%%%%%%%%%%%%%%%%%%%%%%%%%%%%%%%%%%%%%%%%%%%%%%%%%%%%%%

\end{document}